\documentclass[10pt, a4paper]{article}

\usepackage{lrec-coling2024} % this is the new style
\usepackage{enumitem}
\usepackage{multirow}
\usepackage{booktabs}
\usepackage{xcolor}
\usepackage{colortbl}
\usepackage{subfigure}
\usepackage{amsmath}

\setlist[itemize]{itemsep=0pt,parsep=0pt}

\title{Tug-of-War Between Knowledge: Exploring and Resolving Knowledge Conflicts in Retrieval-Augmented Language Models}

\name{Zhuoran Jin$^{1, 2}$, Pengfei Cao$^{1, 2}$, Yubo Chen$^{1, 2}$, Kang Liu$^{1, 2}$, \\ {\bf \large Xiaojian Jiang$^{3}$, Jiexin Xu$^{3}$, Qiuxia Li$^{3}$, Jun Zhao$^{1, 2}$}}

\address{$^1$  School of Artificial Intelligence, University of Chinese Academy of Sciences, Beijing, China\\
$^2$ The Laboratory of Cognition and Decision Intelligence for Complex Systems,\\
Institute of Automation, Chinese Academy of Sciences, Beijing, China \\
$^3$ China Merchants Bank\\
\texttt{\{zhuoran.jin, pengfei.cao, yubo.chen, kliu, jzhao\}@nlpr.ia.ac.cn}\\}

\abstract{
Retrieval-augmented language models (RALMs) have demonstrated significant potential in refining and expanding their \textbf{internal memory} by retrieving evidence from \textbf{external sources}.
However, RALMs will inevitably encounter \textbf{knowledge conflicts} when integrating their internal memory with external sources.
Knowledge conflicts can ensnare RALMs in a tug-of-war between knowledge, limiting their practical applicability.
In this paper, we focus on exploring and resolving knowledge conflicts in RALMs.
First, we present an evaluation framework for assessing knowledge conflicts across various dimensions.
Then, we investigate the behavior and preference of RALMs from the following two perspectives:
(1) Conflicts between internal memory and external sources: We find that stronger RALMs emerge with the \textbf{Dunning-Kruger effect}, persistently favoring their faulty internal memory even when correct evidence is provided.
Besides, RALMs exhibit an \textbf{availability bias} towards common knowledge;
(2) Conflicts between truthful, irrelevant and misleading evidence: We reveal that RALMs follow the principle of \textbf{majority rule}, leaning towards placing trust in evidence that appears more frequently.
Moreover, we find that RALMs exhibit \textbf{confirmation bias}, and are more willing to choose evidence that is consistent with their internal memory.
To solve the challenge of knowledge conflicts, we propose a method called \textbf{C}onflict-\textbf{D}isentangle \textbf{C}ontrastive \textbf{D}ecoding (\textbf{CD\textsuperscript{2}}) to better calibrate the model's confidence.
Experimental results demonstrate that our CD\textsuperscript{2} can effectively resolve knowledge conflicts in RALMs.
 \\ \newline  }

\begin{document}

\maketitleabstract

\section{Introduction}

\begin{figure}[t]
    \centering

    \includegraphics[clip=true,width=0.45\textwidth]{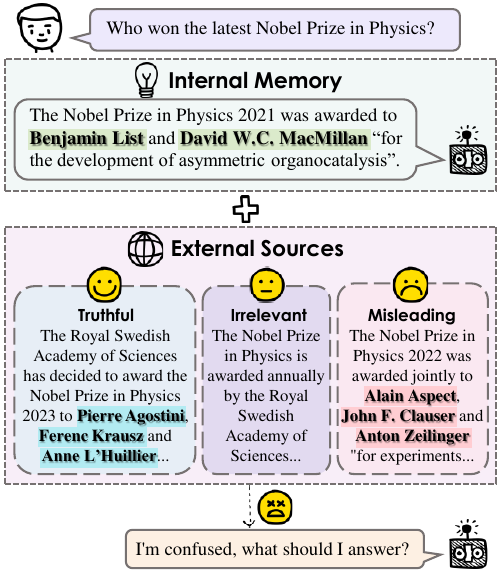}

    % \caption{An example of knowledge conflicts between internal memory and external sources, and between truthful, irrelevant, misleading evidence.}
    \caption{An example of knowledge conflicts.}
    \label{intro}
\end{figure}

% 语言模型在预训练阶段掌握了大量的知识，但是这些知识可能是错误的，过时的，导致语言模型出现幻觉
Large language models (LLMs) \citep{GPT3, GPT4} have memorized a substantial amount of factual knowledge during pre-training, and encapsulated this knowledge within their parameters as \textbf{internal memory} or \textbf{parametric knowledge} \citep{zhu2023physics, sun2023head}.
Nevertheless, the internal memory may sometimes be inaccurate or outdated, making LLMs prone to hallucination \cite{ji2023survey}, where generated responses may seem plausible but are fictional.
As shown in Figure \ref{intro}, for the question ``\textit{Who won the latest Nobel Prize in Physics?}'', its correct answer is ``\textit{Pierre Agostini}''. However, relying solely on unfactual internal memory to answer this question may lead to incorrect answers, such as ``\textit{Benjamin List}''.

% causes the LLM to generate fictitious details when answering the question ``\textit{Who won the latest Nobel Prize in Physics?}''.

To address the issue of hallucination, one promising solution is to employ retrieval-augmented language models (RALMs) \citep{RAG, FiD, ram2023context, shi2023replug, jin2023instructor}.
RALMs first retrieve a handful of reference evidence from \textbf{external sources} or \textbf{non-parametric knowledge}, and subsequently integrate these external sources with their internal memory for generation.
However, owing to challenges like misinformation, divergent perspectives and the evolving nature of knowledge \cite{temporalwiki}, \textbf{knowledge conflicts} \citep{longpre-etal-2021-entity, chen-etal-2022-rich, xie2023adaptive} widely exist in RALMs.
As illustrated in Figure \ref{intro}, the model's internal memory (``\textit{Benjamin List}'') may conflict with external sources (``\textit{Pierre Agostini}''), causing interference.
Moreover, knowledge conflicts also arise among external sources due to the presence of \textbf{truthful} (``\textit{Pierre Agostini}''), \textbf{irrelevant} and \textbf{misleading} (``\textit{Alain Aspect}'') evidence.
\textit{Knowledge conflicts can ensnare RALMs in a tug-of-war between knowledge}, making them confusing and potentially limiting their practical applicability.
% This issue is crucial, while there is a lack of research on it.
Therefore, it is necessary to \textbf{explore} and \textbf{resolve} knowledge conflicts in retrieval-augmented language models.

% 我们为了分析
In this paper, we introduce an evaluation framework that conducts a thorough investigation into knowledge conflicts: (1) We reconstruct four question answering datasets to assess knowledge conflicts from open-domain (\textit{i.e.}, single evidence), entity-centric (\textit{i.e.}, question with entity popularity), and multi-hop (\textit{i.e.}, multiple pieces of evidence) perspectives;
% Baichuan2
(2) We employ seven open-source LLMs (\textit{e.g.}, FLAN-T5, LLaMA2) and one API-access LLM (\textit{e.g.}, ChatGPT) to analyze the influence of model sizes and capabilities under conflicts;
(3) We evaluate RALMs from three key dimensions: \textbf{correctness} (\textit{i.e.}, whether RALMs can answer questions correctly based on internal memory or external sources?), \textbf{faithfulness} (\textit{i.e.}, whether RALMs refer to external sources in answering, and which type of evidence is preferred?) and \textbf{memorization} (\textit{i.e.}, how well RALMs stick to their internal memory?).
% (4) We leverage closed-book QA to induce the internal memory of RALMs, and then distill counterfactuals with ChatGPT to generate high-quality conflicting knowledge automatically.

Based on this, we systematically investigate the behavior and preference of RALMs in the conflicting situation from the following two perspectives:

\textbf{Knowledge Conflicts Between Internal Memory and External Sources}. (1) When we provide knowledge that conflicts with the internal memory of RALMs as external sources, we observe that RALMs \textbf{easily change} their internal beliefs to believe the external knowledge. \textit{As model sizes and capabilities expand, the model gains greater confidence in its internal memory and has a certain ability to perceive conflicts};
% However, if the model is not calibrated well, it will still answer incorrectly;
% naturally, 
(2) Then, we come to wonder \textit{whether the model has different preferences for various types of knowledge?}
We show that RALMs exhibit an \textbf{availability bias} towards common knowledge, \textit{preferring knowledge that is easily accessible in memory}.
% Our finding illustrates that \textit{RALMs exhibit a heightened confirmation bias towards widely accepted knowledge}
For those long-tail knowledge, the model depends on external sources in most cases.
% the model tends to rely more on external sources.
As the knowledge becomes more common, the model gradually shifts towards placing more trust in its internal memory;
(3) Besides, we further conduct an in-depth analysis from the viewpoint of correct and incorrect internal memory.
We find that \textit{capable LLMs may lack confidence in their correct internal memory, but will stubbornly persist in their incorrect internal memory}.
% like LLaMA2 and ChatGPT 
% For example, despite being presented with correct external evidence, ChatGPT tends to maintain confidence in its incorrect internal memory over half the time.
For instance, even when provided with correct external evidence, ChatGPT often persists in trusting its incorrect internal memory for more than half the time.
This phenomenon aligns with the \textbf{Dunning-Kruger effect} in human psychology \citep{kruger1999unskilled, dunning2011dunning, singh2023confidence}, wherein individuals tend to overestimate their abilities when they have limited competence in specific domains.

% (2) Besides, we further conduct an in-depth analysis from the viewpoint of correct and incorrect internal memory.
% We find that \textit{powerful LLMs like LLaMA2 and ChatGPT may lack confidence in their correct internal memory, but will stubbornly persist in their incorrect internal memory}.
% % For example, despite being presented with correct external evidence, ChatGPT tends to maintain confidence in its incorrect internal memory over half the time.
% For instance, even when provided with correct external evidence, ChatGPT often persists in trusting its incorrect internal memory for more than half the time.
% This phenomenon aligns with the \textbf{Dunning-Kruger effect} in human psychology \citep{kruger1999unskilled, dunning2011dunning, singh2023confidence}, wherein individuals with limited competence in a specific domain tend to overestimate their abilities;
% (3) Additionally, we also show that RALMs exhibit a heightened \textbf{availability bias} towards common knowledge, \textit{preferring knowledge that is easily accessible in memory}.
% % Our finding illustrates that \textit{RALMs exhibit a heightened confirmation bias towards widely accepted knowledge}
% For those long-tail knowledge,
% % the model tends to rely more on external sources.
% the model depends on external sources in most cases.
% As the knowledge becomes more common, the model begins to lean towards trusting its internal memory.

\textbf{Knowledge Conflicts Between Truthful, Irrelevant and Misleading Evidence}. (1) RALMs often struggle to discern truthful evidence from misleading evidence, and at times, they may be distracted by irrelevant data.
RALMs follow the principle of \textbf{majority rule}, \textit{leaning towards placing trust in evidence that appears more frequently};
(2) Furthermore, we consider a more interesting scenario where external sources contain evidence supporting the model's internal memory.
To this end, we induce the model's internal memory and integrate it into external sources.
% For example, if the model has incorrect internal memory, then we induce the model's internal memory and add them to external sources as misleading evidence.
% Instead, we induce the model's correct internal memory as truthful evidence.
We find RALMs exhibit \textbf{confirmation bias} \cite{nickerson1998confirmation, xie2023adaptive}, \textit{being more inclined to choose evidence that is consistent with their own internal memory, regardless of whether it is correct or incorrect};
(3) We also examine knowledge conflicts in multi-hop reasoning scenarios, where RALMs must integrate multiple pieces of evidence to answer the question.
\textit{As the count of conflicting hops increases, the model faces greater difficulty in reasoning the correct answer.}

% 难以感知正确和错误
% 随大流，长度
% 推理能力

To solve the challenge of knowledge conflicts, we propose a novel method called \textbf{C}onflict-\textbf{D}isentangle \textbf{C}ontrastive \textbf{D}ecoding (\textbf{CD\textsuperscript{2}}) to better calibrate the model's confidence.
For knowledge conflicts between internal memory and external sources, to mitigate the impact of the RALM's incorrect internal memory, we leverage contrastive decoding to amplify the difference between output logits with and without external sources.
To address knowledge conflicts between truthful, irrelevant and misleading evidence, we adopt \textbf{fact-aware} instruction tuning to enable the RALM to be aware of truthful and misleading evidence.
In detail, we train an \textbf{expert} LM to generate truthful answers and an \textbf{amateur} LM to generate misleading answers, and then utilize contrastive decoding to maximize the difference between expert logits and amateur logits.

Our contributions are summarized as follows:
\begin{itemize}

\item We introduce an evaluation framework that conducts a thorough investigation into knowledge conflicts in RALMs.
We conduct experiments with eight models on four QA datasets, and then evaluate them from three dimensions: correctness, faithfulness and memorization.

\item We investigate the behavior and preference of RALMs in the tug-of-war between knowledge.
We find that powerful RALMs emerge with the Dunning-Kruger effect when internal memory and external sources conflict.
Moreover, RALMs follow the principle of majority rule and exhibit confirmation bias when facing truthful, irrelevant and misleading evidence.

\item We propose a novel method called Conflict-Disentangle Contrastive Decoding (CD\textsuperscript{2}) to better calibrate the model's confidence.
Experimental results demonstrate that our CD\textsuperscript{2} can effectively resolve knowledge conflicts in RALMs, with an average Recall improvement of 2.35\% and 2.41\% on two datasets.
Our code will be publicly available at \url{https://github.com/jinzhuoran/KConflict/}.

\end{itemize}

\section{Related Work}
\subsection{Retrieval-Augmented Language Models}

Retrieval-augmented language models (RALMs) have shown remarkable potential in mitigating hallucinations \cite{chen2023benchmarking} and expanding knowledge boundaries \cite{ren2023investigating}.
% A RALM consists of a parametric language model (LM) and a non-parametric external knowledge store such as Wikipedia, which can effectively solve various knowledge-intensive tasks.
Previous methods \citep{guu2020retrieval, RAG, FiD, borgeaud2022improving} focus on pre-training a smaller LM to better utilize retrieved evidence.
Recently, some work \citep{yu2023generate, ram2023context, shi2023replug} has unleashed the in-context learning capabilities of LLMs, simply prepending retrieved evidence to the input without updating any parameters.
In this paper, we keep in line with the in-context learning paradigm.
Thus, we adopt frozen LLMs to extract answers from a handful of retrieved passages.
% Furthermore, we consider that LMs trained in the pre-training paradigm tend to be relatively small and have limited accurate internal memory.
We do not consider those RALMs like RAG \cite{RAG} and FiD \cite{FiD}, because their limited size makes it difficult to answer questions correctly relying solely on their internal memory.

\subsection{Knowledge Conflicts}
% 让模型做选择不如直接回答
% 我们分析了开原模型，我们有多维度的评价
% 我们从正确错误两个角度
% With the development of retrieval-augmented and tool-augmented, 
Knowledge conflicts \citep{longpre-etal-2021-entity, chen-etal-2022-rich, shi2023trusting,  xie2023adaptive, wang2023resolving, neeman-etal-2023-disentqa} have recently garnered the attention of researchers.
\citet{longpre-etal-2021-entity} propose an entity substitution framework to create entity-based conflicts by replacing a gold entity mention with an alternate entity.
They show that models frequently rely on their parametric knowledge, generating answers not present in the given evidence.
\citet{chen-etal-2022-rich} consider a more realistic scenario, where models consider multiple evidence passages.
Different from \citet{longpre-etal-2021-entity}, they find that when provided with a high recall retriever, models rely almost exclusively on the evidence passages rather than internal memory.
\citet{xie2023adaptive} argue that LLMs do not trust entity substitution-based conflicting evidence, possibly because of the incoherence of the evidence.
To this end, they instruct LLMs to generate coherent conflicting evidence, and reveal that when both supportive and contradictory evidence to their parametric memory are present, LLMs tend to cling to their parametric memory.
However, most of the existing studies only consider the situation where the model encounters conflicts when its internal memory is correct.
In this paper, we separately investigate the behavior of RALMs in cases of both correct and incorrect internal memory and uncover some unique phenomena.
Besides, we also propose an effective method to solve knowledge conflicts in RALMs.

\begin{table*}[h]\centering
\scalebox{0.75}{
\begin{tabular}{l|ccc|ccc|ccc|c}
\toprule
\multirow{2}{*}{\textbf{Model}} & \multicolumn{3}{c|}{\textbf{Internal Memory}} & \multicolumn{3}{c|}{\textbf{Correct Memory w/ C}} & \multicolumn{3}{c|}{\textbf{Incorrect Memory w/ C}} & \multirow{2}{*}{\textbf{IMR - CMR}} \\ 
              & \textbf{EM}    & \textbf{F1}    & \textbf{R}   & \textbf{Mem R} & \textbf{Con R} & \textbf{MR}    & \textbf{Mem R} & \textbf{Con R} & \textbf{MR} &   \\ \midrule
\multicolumn{11}{c}{\textbf{NQ}} \\
FLAN-T5-XL    & 10.34 & 16.68 & 19.02 & 14.42    & 82.87    & 7.20  & 6.41     & 74.75    & 1.32  & -5.88 \\
FLAN-T5-XXL   & 14.43 & 21.29 & 23.44 & 14.36    & 82.53    & 9.83  & 7.86     & 77.63    & 2.83  & -7.00 \\
FLAN-UL2      & 22.92 & 29.69 & 30.11 & 11.33    & \textbf{86.13}    & 7.11  & 10.22    & \textbf{76.40}    & 4.75  & -2.36 \\
Baichuan2 7B  & 21.88 & 30.18 & 32.47 & 16.51    & 83.40    & 10.31 & 16.82    & 65.71    & 10.20  & -0.11 \\
Baichuan2 13B & 26.02 & 34.40 & 37.33 & 22.34    & 77.31    & 15.91 & 21.19    & 58.68    & 12.76  & -3.15 \\
LLaMA2 7B     & 29.04 & 39.24 & 41.72 & 19.04    & 82.03    & 13.10 & 22.95    & 65.77    & 12.92 & -0.18 \\
LLaMA2 13B    & \textbf{34.86} & \textbf{46.27} & 50.26 & 15.30    & 84.90    & 8.47  & 26.02    & 60.27    & 17.53 & 9.06 \\
ChatGPT       & 25.77 & 38.56 & \textbf{59.84} & \textbf{27.68}    & 80.40    & \textbf{18.14} & \textbf{39.23}    & 62.58    & \textbf{50.69} & \textbf{32.55} \\ \midrule
\multicolumn{11}{c}{\textbf{TriviaQA}} \\
\multicolumn{1}{l|}{FLAN-T5-XL} &
  27.43 &
  32.76 &
  \multicolumn{1}{c|}{33.83} &
  16.74 &
  76.94 &
  \multicolumn{1}{c|}{16.10} &
  8.97 &
  84.37 &
  \multicolumn{1}{c|}{8.84} & -7.26
   \\
\multicolumn{1}{l|}{FLAN-T5-XXL} &
  35.67 &
  41.14 &
  \multicolumn{1}{c|}{41.89} &
  17.89 &
  80.14 &
  \multicolumn{1}{c|}{17.85} &
  8.88 &
  83.96 &
  \multicolumn{1}{c|}{8.54} & -9.31
   \\
\multicolumn{1}{l|}{FLAN-UL2} &
  47.47 &
  53.42 &
  \multicolumn{1}{c|}{54.82} &
  16.10 &
  82.86 &
  \multicolumn{1}{c|}{15.92} &
  19.50 &
  \textbf{84.60} &
  \multicolumn{1}{c|}{19.70} & 3.78
   \\
\multicolumn{1}{l|}{Baichuan2 7B} &
  56.58 &
  60.71 &
  \multicolumn{1}{c|}{62.32} &
  16.76 &
  \textbf{82.30} &
  \multicolumn{1}{c|}{15.96} &
  18.55 &
  76.03 &
  \multicolumn{1}{c|}{15.90} & -0.06
   \\
\multicolumn{1}{l|}{Baichuan2 13B} &
  60.10 &
  64.49 &
  \multicolumn{1}{c|}{66.63} &
  22.46 &
  78.89 &
  \multicolumn{1}{c|}{\textbf{21.27}} &
  19.54 &
  73.71 &
  \multicolumn{1}{c|}{18.94} & -2.33
   \\
\multicolumn{1}{l|}{LLaMA2 7B} &
  61.58 &
  65.46 &
  \multicolumn{1}{c|}{66.60} &
  19.16 &
  79.14 &
  \multicolumn{1}{c|}{19.10} &
  20.33 &
  76.67 &
  \multicolumn{1}{c|}{17.39} & -1.71
   \\
\multicolumn{1}{l|}{LLaMA2 13B} &
  \textbf{67.20} &
  \textbf{72.01} &
  \multicolumn{1}{c|}{74.04} &
  16.76 &
  81.90 &
  \multicolumn{1}{c|}{16.16} &
  18.87 &
  75.14 &
  \multicolumn{1}{c|}{18.39} & 2.23
   \\
\multicolumn{1}{l|}{ChatGPT} &
  56.60 &
  66.30 &
  \multicolumn{1}{c|}{\textbf{82.88}} &
  \textbf{26.84} &
  76.17 &
  \multicolumn{1}{c|}{19.57} &
  \textbf{39.72} &
  69.87 &
  \multicolumn{1}{c|}{\textbf{39.41}} & \textbf{19.84}\\
\bottomrule
\end{tabular}
}
\caption{Experimental results under knowledge conflicts between internal memory and external sources on NQ and TriviaQA.
% R denotes the Recall between internal memory predictions and gold references. 
Mem R denotes the Recall between predictions and internal memory predictions. Con R denotes the Recall between predictions and conflicting references. IMR - CMR denotes the difference in memorization ratio between incorrect memory and correct memory. Bold denotes the highest results.}
% $h$ denotes the number of hops.
\label{nq}
\end{table*}

\section{Evaluation Framework}

\subsection{Datasets}

Following previous work \citep{chen-etal-2022-rich, xie2023adaptive}, we adopt the question answering (QA) task to explore knowledge conflicts in the open-book setting.
We choose four QA datasets: NQ \citelanguageresource{NQ}, TriviaQA \citelanguageresource{TQA}, PopQA \citelanguageresource{popqa} and MuSiQue \citelanguageresource{trivedi2022musique}.
NQ and TriviaQA are both commonly used open-domain QA datasets.
PopQA is an entity-centric QA dataset including truly long-tail distribution.
PopQA is constructed from triples in Wikidata where the subject entity in each question has popularity.
MuSiQue is a multi-hop QA dataset with 2-4 hop questions.
 % for analyzing the impact of knowledge conflicts on reasoning
For each dataset, we randomly sample 500 or 1000 questions from the original testing set as the evaluation data.
We only select questions for which supporting evidence can be retrieved.
All the datasets use Wikipedia \citelanguageresource{KILT} as their external source.

\subsection{Analyzed Models}
To analyze the impact of model sizes and capabilities, we adopt seven open-source LLMs, including FLAN-T5-XL 3B \cite{FLAN}, FLAN-T5-XXL 11B, FLAN-UL2 20B \cite{UL2}, Baichuan2 7B \cite{yang2023baichuan}, Baichuan2 13B, LLaMA2 7B \cite{touvron2023llama} and LLaMA2 13B.
Besides, we use the \texttt{gpt-3.5-turbo} version of ChatGPT as the API-access LLM.
% We do not choose those pre-trained retrieval-augmented models, because their small number of parameters makes it difficult to answer questions correctly relying solely on their internal memory.
We use in-context learning to let RALMs answer the questions conditioned on $M \in \{4,8,16\}$ demonstrations and $K \in \{3,5,10,20\}$ retrieved evidence.

\subsection{Evaluation Metrics}
% We evaluate RALMs from three dimensions:
\paragraph{Correctness} According to the previous study, we use EM and F1 to evaluate whether the model can provide correct answers. Since LLMs tend to produce verbose yet accurate responses, which may lead to lower EM and F1 scores. Therefore, following \citet{adlakha2023evaluating}, we adopt Recall (R) to compute the proportion of tokens in the gold reference that are present in the prediction.

% 答案对不对
\paragraph{Faithfulness}
To assess the extent to which the model's predictions depend on retrieved evidence, we adopt K-Precision (KP) \cite{adlakha2023evaluating} to calculate the proportion of tokens in the prediction that are present in external evidence.
K-Precision can also be used to quantify the preference for truthful, irrelevant and misleading evidence.

% 答案从哪来
\paragraph{Memorization}
% 是否坚持原有记忆
We leverage the memorization ratio \cite{longpre-etal-2021-entity, chen-etal-2022-rich, xie2023adaptive} $MR=\frac{f_{m}}{f_{m}+f_{s}}$ to measure how often the model sticks to its internal memory, where $f_{m}$ is the frequency of relying on internal memory, $f_{s}$ is the frequency of relying on external sources.

% 用了两个传统指标，然后用了召回率，

\subsection{Memory Induction and Conflict Generation}
We adopt closed-book QA to induce the model's internal memory and perform greedy decoding of the answers.
Then we prompt the model to further generate evidence supporting its internal memory based on the elicited answers.
To construct more confusing and coherent conflicting knowledge, we distill counterfactuals \citep{xie2023adaptive, chen-etal-2023-disco} with LLMs.
We unleash the creative power of ChatGPT with a temperature $\tau = 1.0$, enabling it to create counterfactual answers and conflicting evidence based on the given questions, reference answers, and supporting evidence.

% \usepackage{multirow}

% % 放一个prompt
% Step 1 in Figure 2 illustrates how we elicit parametric memory: in a closed-book QA fashion, LLMs recall their parametric memory to answer questions without any external evidence.
% Step 1 in Figure 2 illustrates how we elicit parametric memory: in a closed-book QA fashion, LLMs recall their parametric memory to answer questions without any external evidence.

% 不给证据，让模型直接回答答案作为内部记忆，如果正确，我们认为内部记忆也正确。然后让模型生成文章，作为内部记忆的诱导

% 然后

\section{Conflicts Between Internal Memory and External Sources}
\label{section4}

In this section, we investigate knowledge conflicts between internal memory and external sources.
Specifically, we first elicit the model's internal memory through closed-book QA, then classify the internal memory according to whether it is correct or not.
For questions that the model's internal memory can answer correctly/incorrectly, we provide $K=3$ conflicting incorrect/correct external evidence.

\begin{figure*}[t]
    \centering

    \includegraphics[clip=true,width=0.91\textwidth]{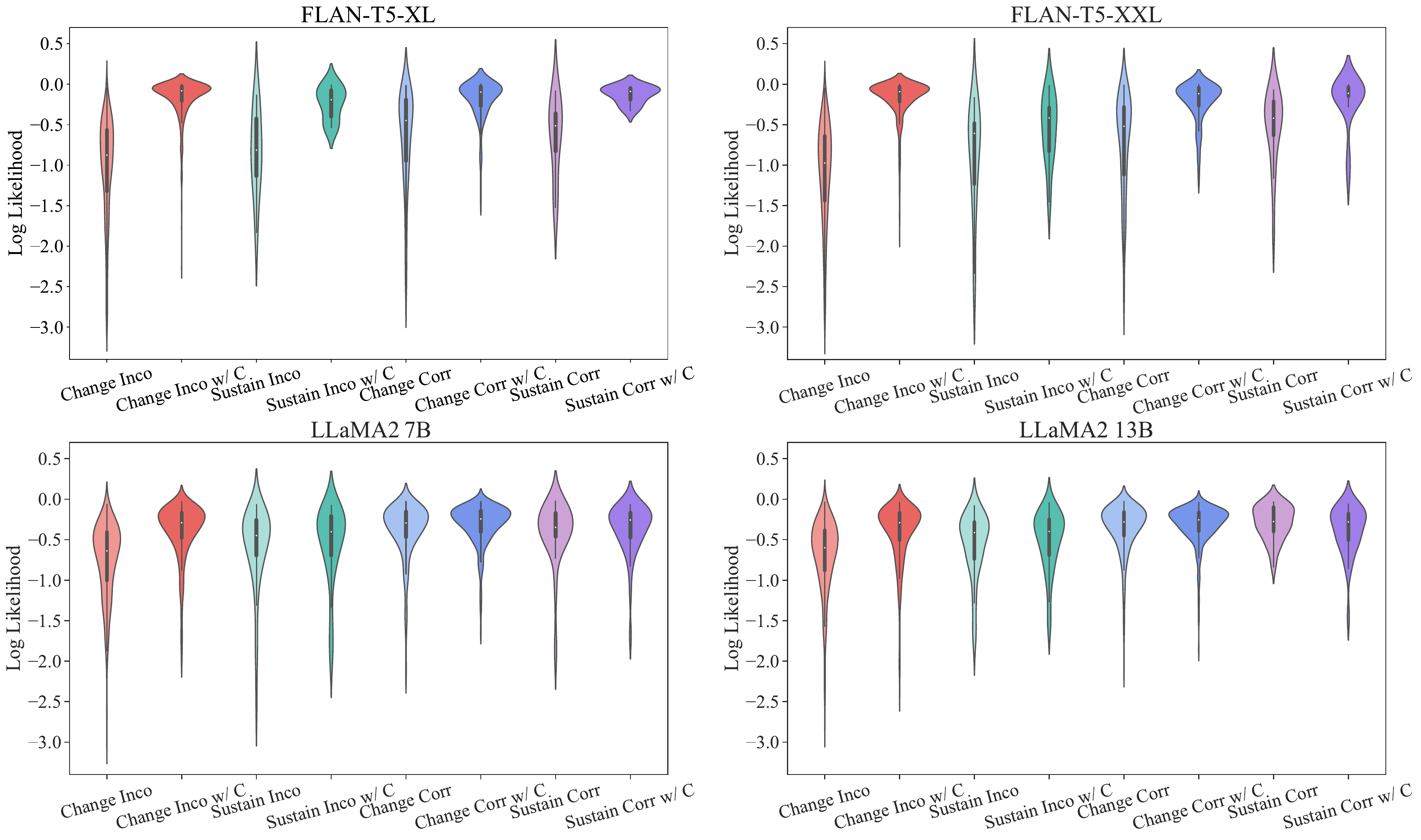}

    \caption{Confidence score distributions of different models under knowledge conflicts between internal memory and external sources on NQ. w/ C denotes providing conflicting evidence to the model. The wider the violin plot, the denser the data points. The larger the log likelihood, the higher the confidence score.}
    \label{violin}
\end{figure*}

\begin{figure*}[h]
    \centering

    \includegraphics[clip=true,width=0.92\textwidth]{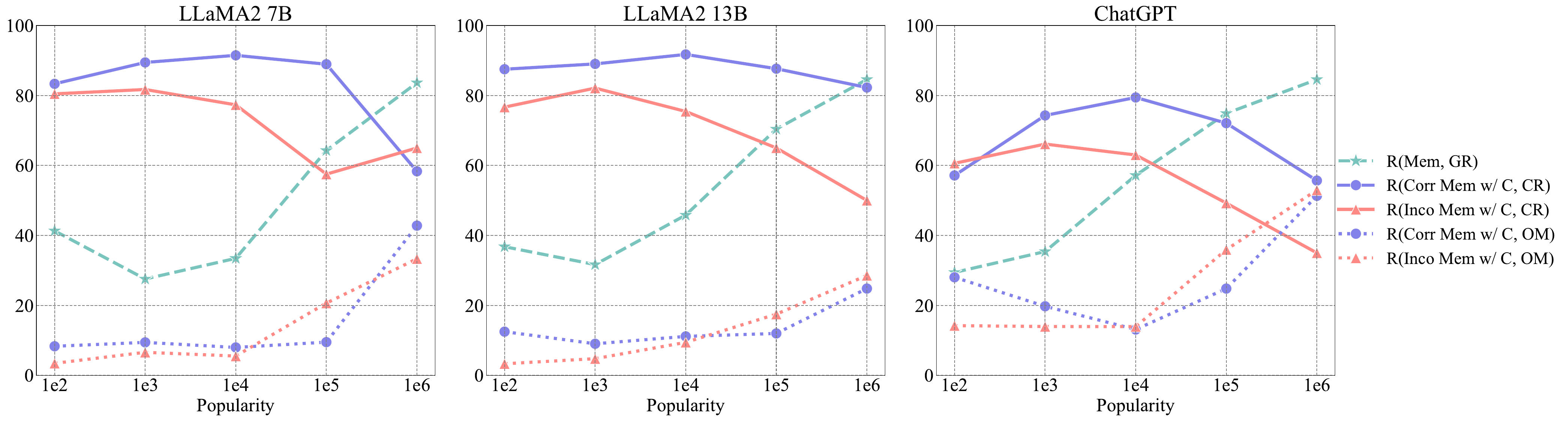}

    \caption{Recall under knowledge conflicts with various entity popularities on PopQA. GR denotes the gold references, CR denotes the conflicting references, and OM denotes the internal memory predictions.}
    \label{pop}
\end{figure*}

% 具体来说，我们先让模型直接回答问题，然后根据模型回答正确与否，将问题分为两类。
% 对于模型正确回答的问题，我们提供冲突的错误知识。
% 对于模型错误回答的问题，我们提供冲突的正确知识。

% 模型

\paragraph{RALMs easily change their internal memory to believe the conflicting external knowledge.}
As shown in Table \ref{nq}, we conduct experiments on NQ and TriviaQA.
We can find that after providing conflicting evidence, Con R (Recall of conflicting references) is much higher than Mem R (Recall of internal memory).
This demonstrates that \textit{RALMs tend to trust external knowledge, even when it conflicts with their initial internal memory.}
% To
% 小模型没有冲突感知能力，大模型有一些

\paragraph{The greater the model's capability, the more confident it becomes.}
Intuitively, we can observe that as the model's capacity increases, the correctness (\textit{e.g.}, EM, F1 and R) of its internal memory also improves.
Meanwhile, we find that along with the increase in correctness, there is also an improvement in the MR.
This indicates that \textit{as model sizes and capabilities expand, the model gains greater confidence in its internal memory}.
To conduct a comprehensive analysis of the changes in the model's confidence score, we categorize the model's behavior under conflict into four distinct groups: (1) Change Inco: Modifying incorrect internal memory based on conflicting evidence; (2) Sustain Inco: Preserving incorrect internal memory despite conflicting evidence; (3) Change Corr: Modifying correct internal memory based on conflicting evidence; (4) Sustain Corr: Preserving correct internal memory despite conflicting evidence.
As shown in Figure \ref{violin}, FLAN-T5 consistently increases its confidence score when encountering conflicts, indicating that FLAN-T5 lacks confidence in internal beliefs and prefers to trust external evidence.
On the contrary, we observe that LLaMA2 significantly improves confidence scores only when modifying its incorrect internal memory based on truthful evidence.
This proves that as the model's capabilities improve, \textit{the model has a certain ability to perceive conflicts and will not blindly trust conflicting evidence}.
However, if the model is not calibrated well to resolve conflicts, it will still answer incorrectly.

\begin{table*}[t]\centering
\scalebox{0.76}{
\begin{tabular}{lccccccccc}
\toprule
\textbf{Model} & \textbf{EM} & \textbf{F1} & \textbf{R} & \textbf{Con R} $\downarrow$
 & \textbf{Tru KP} & \textbf{Mis KP} $\downarrow$
 & \textbf{Irr KP} $\downarrow$
 & \textbf{Corr MR} & \textbf{Inco MR} \\ \midrule
FLAN-T5-XL         & 52.55 & 62.81 & 66.95 & -     & 84.00 & -     & 62.70 & 62.81 & 3.45  \\
\rowcolor[RGB]{230, 230, 230}\ \ \ \ w/ Conflict    & 30.23 & 39.84 & 43.98 & 49.70 & 62.00 & 68.50 & 33.01 & 45.45 & 3.32  \\
FLAN-T5-XXL        & 52.30 & 63.83 & 70.00 & -     & \underline{85.86} & -     & 61.34 & 65.24 & 4.19  \\
\rowcolor[RGB]{230, 230, 230}\ \ \ \ w/ Conflict   & 29.34 & 39.07 & 43.40 & 52.54 & 61.22 & 70.72 & 31.83 & 47.56 & 4.84  \\
FLAN-UL2           & \underline{58.55} & \underline{68.42} & 71.04 & -     & 85.66 & -     & \underline{60.23} & 75.11 & 3.81  \\
\rowcolor[RGB]{230, 230, 230}\ \ \ \ w/ Conflict      & \textbf{31.51} & 39.35 & 41.76 & 53.12 & 58.34 & 71.40 & \textbf{30.71} & 48.93 & 5.99  \\
Baichuan2 7B       & 43.75 & 55.49 & 60.26 & -     & 79.70 & -     & 62.57 & 67.14 & 14.88 \\
\rowcolor[RGB]{230, 230, 230}\ \ \ \ w/ Conflict  & 30.87 & \textbf{41.25} & 45.71 & \textbf{44.32} & 64.61 & \textbf{62.47} & 35.88 & 51.07 & 12.50 \\
Baichuan2 13B      & 48.09 & 59.81 & 64.96 & -     & 81.73 & -     & 60.30 & 72.32 & 15.18 \\
\rowcolor[RGB]{230, 230, 230}\ \ \ \ w/ Conflict & 27.68 & 37.28 & 42.60 & 47.75 & 61.33 & 65.91 & 33.19 & 46.73 & 14.06 \\
LLaMA2 7B          & 48.21 & 59.00 & 61.94 & -     & 80.50 & -     & 62.29 & 69.78 & 16.20 \\
\rowcolor[RGB]{230, 230, 230}\ \ \ \ w/ Conflict     & 30.36 & 39.60 & 42.59 & 49.60 & 61.40 & 66.43 & 34.25 & 49.84 & 14.90 \\
LLaMA2 13B         & 45.28 & 56.79 & 62.99 & -     & 82.42 & -     & 62.78 & 71.10 & 19.59 \\
\rowcolor[RGB]{230, 230, 230}\ \ \ \ w/ Conflict    & 29.85 & 39.93 & 46.14 & 45.73 & 65.85 & 65.41 & 35.73 & 50.13 & 16.28 \\
ChatGPT            & 20.66 & 36.93 & \underline{72.00} & -     & 78.98 & -     & 69.94 & \underline{84.96} & \underline{60.00} \\
\rowcolor[RGB]{230, 230, 230}\ \ \ \ w/ Conflict       & 16.58 & 31.07 & \textbf{61.05} & 46.65 & \textbf{70.82} & 67.75 & 43.41 & \textbf{69.21} & \textbf{51.78} \\ \bottomrule
\end{tabular}
}
\caption{Experimental results under knowledge conflicts between truthful, irrelevant and misleading evidence on NQ. Con R is the Recall between the predictions and conflicting references. Tru KP, Mis KP and Irr KP represent the K-Precision of truthful, misleading and irrelevant evidence. Underline means the highest or lowest ($\downarrow$) results without conflicts. Bold means the highest or lowest ($\downarrow$) results under conflicts.}
% $h$ denotes the number of hops.
\label{wconf}
\end{table*}

\begin{figure*}[h]
    \centering

    \includegraphics[clip=true,width=0.96\textwidth]{bar.pdf}

    \caption{Recall of LLaMA2 7B with various amounts of evidence and different conflicting ratios on NQ.}
    \label{bar}
\end{figure*}

% We can also observe a general improvement in the model's confidence scores, regardless of whether conflicts altered the internal memory or not in Figure \ref{violin}.

\paragraph{The Dunning-Kruger effect in human psychology merges in capable LLMs.}
% are surprised to 
Based on the above findings, a natural question arises: \textit{is the model more confident on its correct internal memory or its incorrect internal memory?}
As shown in Table \ref{nq}, we are surprised to discover that with the increasing capabilities of the model, the memorization ratio for incorrect beliefs gradually surpasses that of correct beliefs.
For instance, less capable LLMs like FLAN-T5-XL and FLAN-T5-XXL seldom adhere to their incorrect memory, whereas more capable LLMs like ChatGPT frequently rely on their incorrect internal memory for over half of the time.

We argue that \textit{there is a phenomenon emerging among those \textbf{capable} models: they may lack confidence in their correct internal memory, but will stubbornly persist in their incorrect internal memory.}
This phenomenon aligns with the \textbf{Dunning-Kruger effect} in human psychology \citep{kruger1999unskilled, dunning2011dunning, singh2023confidence}, wherein individuals with limited competence in a specific domain tend to overestimate their abilities.
As depicted in Figure \ref{violin}, when the model preserves incorrect memory despite conflicting evidence, there are notably high confidence scores present.
This also proves that the model is overconfident in some incorrect internal memory \cite{slobodkin2023curious}.

% 对于错误记忆的信心逐渐超过对于正确记忆的信心
% 模型对于一些错误记忆深刻

% We divide the model's behavior under conflict into four categories:
%  half the time. This phe-
% nomenon is consistent with the Dunning-Kruger
% effect in human psychology;

% LLMs do not trust entity substitution-based counter-memory.

\paragraph{RALMs exhibit an availability bias towards common knowledge.}
We aim to explore whether the model's preference varies when faced with knowledge conflicts of different popularities.
We conduct experiments on entity-centric PopQA with a knowledge popularity distribution ranging from 1e2 to 1e6.
As depicted in Figure \ref{pop}, for those long-tail knowledge,
% the model tends to rely more on external sources.
the model depends on external sources in most cases.
However, as the knowledge popularity increases, the model begins to lean towards trusting its internal memory.
% At the same time, for those common knowledge, the model leans towards trusting its internal memory.
This indicates when facing conflicts, the model exhibits an availability bias towards commonly known knowledge, more willing to believe in knowledge that they can easily recall.

% (3) Additionally, we also show that RALMs exhibit a heightened \textbf{availability bias} towards common knowledge.
% Our finding illustrates that \textit{RALMs exhibit a heightened confirmation bias towards widely accepted knowledge}
% For those long-tail knowledge, the model tends to rely more on external sources.
% At the same time, \textit{for those common knowledge, the model leans towards trusting its internal memory}.

% In this section, We experiment LLMs in the single-source evidence setting where counter-memory is the sole evidence presented to LLMs. Such knowledge conflict happens when LLMs are augmented with tools returning single external evidence such as Wikipedia API (Yao et al., 2023). In particular, for counter-memory construction, we would apply 1) the entity substitution counter-memory method, a widely-applied strategy in previous work, and 2) our generation-based metho

\section{Conflicts Between Truthful, Irrelevant and Misleading Evidence}
\label{section5}
In this section, we investigate knowledge conflicts between truthful, irrelevant and misleading evidence.
First, we provide the model with $K=10$ evidence (including truthful and irrelevant evidence) and then prompt it to perform open-book QA.
Then, we construct conflicts by introducing misleading evidence, ensuring that the number of misleading evidence equals that of truthful evidence.

\paragraph{RALMs often
struggle to discern truthful evidence from misleading evidence.}
As shown in Table \ref{wconf}, after encountering conflicts, the correctness of RALMs will decline to a certain extent.
Among them, the Recall of FLAN-UL2 drops the most ($\downarrow$ 29.28\%), and the Recall of ChatGPT drops the least ($\downarrow$ 10.95\%).
This also confirms our previous argument that the more capable model possesses a certain ability to perceive conflicts.
However, for most RALMs, their K-Precision for misleading evidence is higher than that for truthful evidence, indicating that \textit{they still have difficulty distinguishing truthful evidence from misleading evidence.}
Furthermore, we also observe that they can be easily distracted by irrelevant evidence.

\paragraph{RALMs follow the principle of majority rule.}
We aim to delve deeper into the criteria the model employs when selecting evidence as the reference for generating answers.
We provide LLaMA2 7B with different amounts of evidence ($K \in \{5,10,20\}$) and set different conflicting ratios (truthful evidence: misleading evidence $\in \{2:0, 2:1, 2:2, 1:2, 0:2\}$).
As depicted in Figure \ref{bar}, the model leans towards placing trust in evidence that appears more frequently.
Besides, we find that \textit{more external evidence is not necessarily better}.
Excessively long context may cause the model to lose focus, making it challenging to provide accurate answers.

\begin{figure}[h]
    \centering

    \includegraphics[clip=true,width=0.49\textwidth]{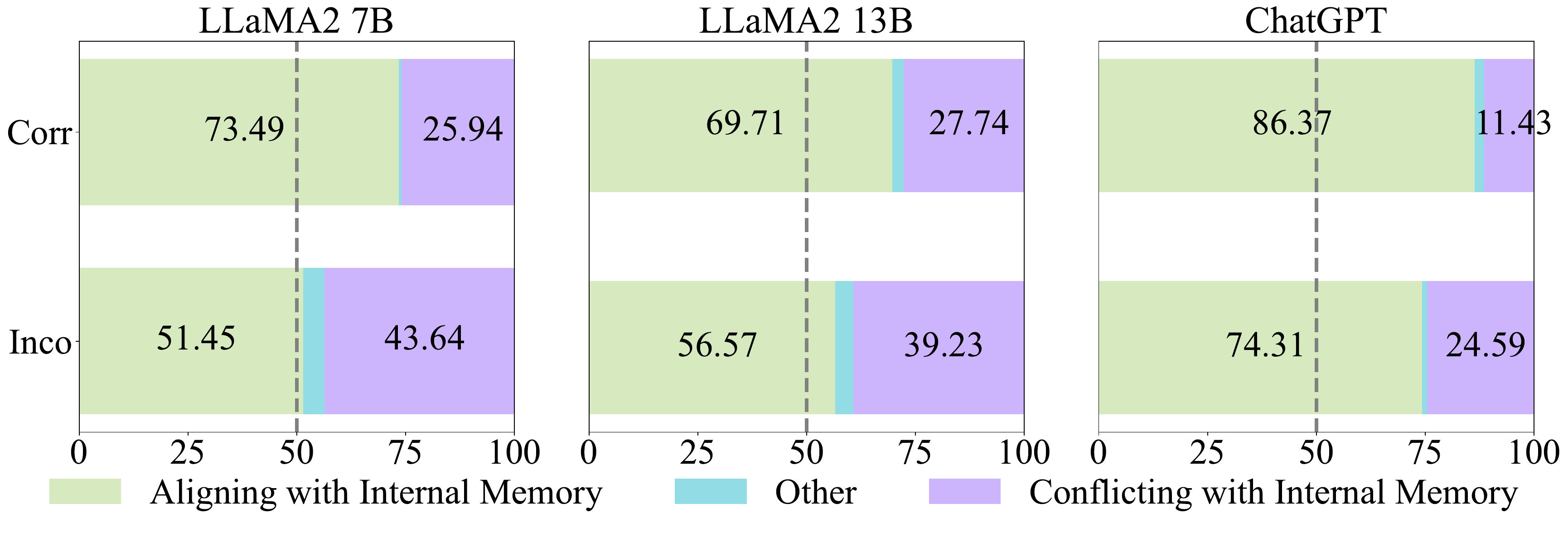}

    % \caption{An example of knowledge conflicts between internal memory and external sources, and between truthful, irrelevant, misleading evidence.}
    \caption{Frequency of choosing evidence aligning with or conflicting with internal memory on NQ.}
    \label{mem}
\end{figure}
\vspace{-10pt}

\paragraph{RALMs exhibit confirmation bias when partial external evidence aligns with their internal memory.}
Furthermore, we consider a more interesting scenario if external sources contain evidence supporting the model's internal memory.
For example, if the model has incorrect internal memory, then we induce the model's internal memory and add it to external sources as misleading evidence.
Instead, we induce the model's correct internal memory as truthful evidence.
As depicted in Figure \ref{mem}, RALMs exhibit \textbf{confirmation bias} \cite{nickerson1998confirmation}, \textit{displaying a stronger inclination to select evidence that aligns with their own internal memory.}
This is consistent with the observations of \citet{xie2023adaptive}.
Besides, we also find that \textit{as model sizes and capabilities expand, the model gains greater confidence in its internal memory, especially when the internal memory is provided as external evidence}, which further supports our findings in Section \ref{section4}.

\begin{figure}[h]
    \centering

    \includegraphics[clip=true,width=0.46\textwidth]{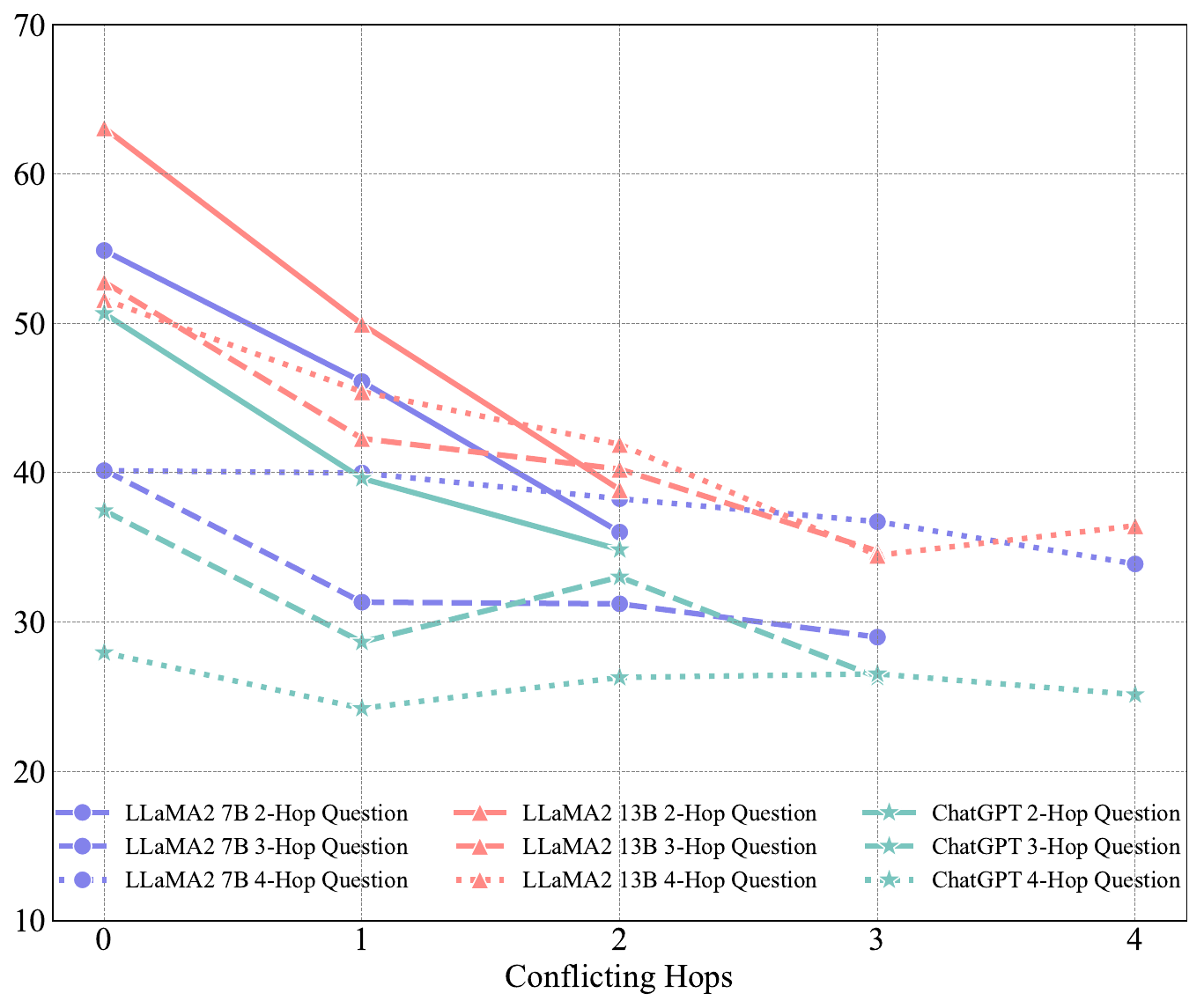}

    % \caption{An example of knowledge conflicts between internal memory and external sources, and between truthful, irrelevant, misleading evidence.}
    \caption{Recall of different models with various conflicting hops on MuSiQue.}
    \label{hop}
\end{figure}

\paragraph{As the number of conflicting hops increases, the model faces greater difficulty in reasoning.}
To investigate knowledge conflicts in multi-hop reasoning scenarios, we conduct experiments on MuSiQue.
For multi-hop questions, the model needs to reason based on multiple pieces of evidence.
Therefore, we incrementally raise the count of misleading evidence until it matches the count of truthful evidence.
As illustrated in Figure \ref{hop}, it is evident that  \textit{an increase in the number of conflicting hops hinders the model's ability to reason accurately.}
Moreover, we also find that ChatGPT does not perform well in scenarios involving multi-hop reasoning with multiple pieces of evidence.

\section{Conflict-Disentangle Contrastive Decoding}

\begin{figure}[t]
    \centering

    \includegraphics[clip=true,width=0.46\textwidth]{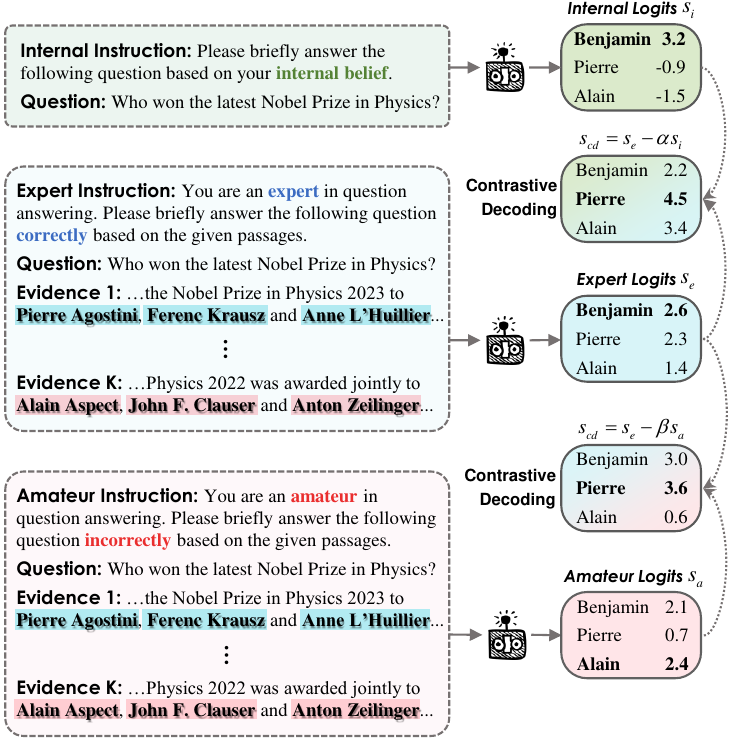}

    \caption{An illustration of Conflict-Disentangle Contrastive Decoding method.}
    \label{method}
\end{figure}

Based on our investigations above, we believe that a capable RALM has a certain ability to \textbf{perceive} conflicts.
However, the model still struggles to answer correctly as it lacks proper calibration to \textbf{resolve} conflicts.
As shown in Figure \ref{method}, after providing the external sources, the logit of token ``\textit{Pierre}'' increases significantly, while the logit of token ``\textit{Benjamin}'' decreases.
Although the knowledge conflict does lead to a shift in the model's confidence, it still tends to produce an inaccurate answer because of the strong influence of its incorrect internal memory.
Therefore, we focus on better calibrating the RALM's confidence to resolve knowledge conflicts.

\subsection{Method}

\begin{table*}[t]\centering
\scalebox{0.78}{
\begin{tabular}{l|cccccc|cccccc}
\toprule
\multirow{2}{*}{\textbf{Method}} & \multicolumn{6}{c|}{\textbf{NQ-Conf}}                                  & \multicolumn{6}{c}{\textbf{TriviaQA-Conf}}                             \\
                        & \textbf{EM}    & \textbf{F1}    & \textbf{R} & \textbf{Tru KP} & \textbf{Mis KP} $\downarrow$ & \textbf{Irr KP} $\downarrow$ & \textbf{EM}    & \textbf{F1}    &  \textbf{R}  & \textbf{Tru KP} & \textbf{Mis KP} $\downarrow$& \textbf{Irr KP} $\downarrow$\\ \midrule
Closed-book & 31.38 & 41.86 & 44.88 & - & - & - & 65.42 & 68.89 & 70.14 & - & - & - \\
In-context     & 30.36 & 39.60 & 42.59 &  61.40  & 66.43  & 34.25  & 55.38 & 59.57 & 60.87 & 73.36  & 54.14  & 19.65  \\
Finetune\textsubscript{Orig}    & 46.05 & 54.34 & 55.50 & 72.68  & 55.36  & 33.29  & 64.56 & 69.96 & 71.54 & 82.87  & 45.81  & \textbf{19.61}  \\
Finetune\textsubscript{Conf}     & 58.67 & 68.17 & 69.77 & 83.06  & 40.04  & 33.51  & 74.18 & 80.19 & 81.81  & 91.67  & 32.74  & 23.29  \\
CD\textsuperscript{2} $\beta=0.5$                     & \textbf{61.35} & \textbf{70.83} & \textbf{72.42} & \textbf{85.15}  & \textbf{39.62}  & \textbf{32.55}  & \textbf{77.76} & \textbf{82.47} & \textbf{83.93} & \textbf{93.03}  & \textbf{31.60}  & 23.51  \\ \bottomrule
\end{tabular}
}
\caption{Performance of LLaMA2 7B on NQ-Conf and TriviaQA-Conf.}
\label{conf2}
\end{table*}

Contrastive decoding \cite{li-etal-2023-contrastive, shi2023trusting}, a search objective for generating fluent and diverse text, aims to search for the higher-quality output token that maximizes the difference between expert (positive) logits and amateur (negative) logits.
Inspired by this, we propose a novel method called \textbf{C}onflict-\textbf{D}isentangle \textbf{C}ontrastive \textbf{D}ecoding (\textbf{CD\textsuperscript{2}}) to maximize the difference between various logits under knowledge conflicts and calibrate the model's confidence shown in Figure \ref{method}.

For knowledge conflicts between internal memory and external sources, to mitigate the impact of the RALM's incorrect internal memory, we predict the output token $y_{t}$ by amplifying the difference between expert logits with external sources $s_{e}$ and internal logits without external sources $s_{i}$:
\begin{equation*}
y_{t} = \mathrm{argmax}\left(s_{e}\left(y_{t}\mid d,x,y_{<t}\right)-\alpha s_{i}\left(y_{t}\mid x,y_{<t}\right)\right),
\end{equation*}
where $d$ denotes the external sources, $x$ denotes the question, $y$ denotes the answer.
Our method can be applied to LLMs like LLaMA2 during the inference stage without additional training.

For knowledge conflicts between truthful, irrelevant and misleading evidence, we first adopt \textbf{fact-aware} instruction tuning to enable the RALM to be aware of truthful and misleading evidence in retrieved external sources.
As shown in Figure \ref{method}, we train an \textbf{expert} LM to generate truthful answers normally, and an \textbf{amateur} LM to generate misleading answers deliberately.
Then, we predict the output token $y_{t}$ by maximizing the difference between expert logits $s_{e}$ and amateur logits $s_{a}$:
\begin{equation*}
y_{t} = \mathrm{argmax}\left(s_{e}\left(y_{t}\mid d,x,y_{<t}\right)-\beta s_{a}\left(y_{t}\mid d,x,y_{<t}\right)\right).
\end{equation*}

% \usepackage{multirow}

\subsection{Experiment Setups}

We employ LLaMA2 7B as the backbone model.
We conduct experiments on two widely used QA datasets: NQ and TriviaQA.
For knowledge conflicts between internal memory and external sources, we create NQ-Inco and TriviaQA-Inco datasets, including questions that the model's internal memory answers incorrectly.
For knowledge conflicts between truthful, irrelevant and misleading evidence, we reconstruct NQ-Conf and TriviaQA-Conf datasets following the setting in Section \ref{section5}. 
We provide the model with $K=10$ evidence.
Then we compare our CD\textsuperscript{2} with the following baselines: (1) Closed book: We do not provide external sources and directly let the model answer the question; (2) In-context: We let the model answer the question based on external sources; (3) Finetune\textsubscript{Orig}: We finetune the model on the original training set; (4) Finetune\textsubscript{Conf}: We finetune the model on the conflict-enhanced training set (\textit{i.e.}, knowledge conflicts exist in the external sources).
Our CD\textsuperscript{2} also adopts fact-aware instruction tuning on the conflict-enhanced training set.
For all methods that require training, we apply LoRA \cite{hu2022lora} to finetune the model on NQ with 3,000 training examples.
We set learning rate=5e-5, epoch=5, batch size=4, lora rank=8 and lora alpha=16.
All experiments are conducted with NVIDIA RTX A6000 GPUs.

\subsection{Results}

On the one hand, as shown in Table \ref{conf1}, CD\textsuperscript{2} can be directly applied to LLaMA2 7B without updating model parameters.
We can observe that CD\textsuperscript{2} leads to a 2.05\% and 2.70\% Recall improvement on NQ-Inco and TriviaQA-Inco respectively.
This highlights the efficacy of our method in mitigating the model's incorrect internal memory and enhancing its attention towards external sources.
On the other hand, as shown in Table \ref{conf2}, CD\textsuperscript{2} enables the model to effectively distinguish truthful evidence from misleading evidence.
Compared with In-context, CD\textsuperscript{2} achieves a notable 29.83\% and 23.06\% Recall improvement on NQ-Conf and TriviaQA-Conf, while Mis KP decreases significantly.

\begin{table}[t]\centering
\scalebox{0.82}{
\begin{tabular}{llccc}
\toprule
\textbf{Dataset}          & \textbf{Method} & \textbf{EM}    & \textbf{F1}    & \textbf{R}     \\ \midrule
\multirow{4}{*}{\textbf{NQ-Inco}}       & In-context        & 31.55          & 41.87          & 43.63          \\
                          & \ \ + CD\textsuperscript{2} $\alpha=0.3$           & \textbf{32.29} & \textbf{43.07} & 44.98          \\
                          & \ \ + CD\textsuperscript{2} $\alpha=0.5$          & 32.10          & 43.04          & \textbf{45.68} \\
                          & \ \ + CD\textsuperscript{2} $\alpha=0.7$          & 31.73          & 42.19          & 44.36          \\ \midrule
\multirow{4}{*}{\textbf{TriviaQA-Inco}} & In-context        & 51.38          & 55.57          & 56.61          \\
                          & \ \ + CD\textsuperscript{2} $\alpha=0.3$          & \textbf{52.41} & 57.46          & 58.97          \\
                          & \ \ + CD\textsuperscript{2} $\alpha=0.5$          & \textbf{52.41} & \textbf{57.72} & \textbf{59.31} \\
                          & \ \ + CD\textsuperscript{2} $\alpha=0.7$          & 50.34          & 56.13          & 58.07          \\ \bottomrule
\end{tabular}
}
\caption{Performance of LLaMA2 7B on NQ-Inco and TriviaQA-Inco.}
\label{conf1}

\end{table}

\section{Conclusion}

In this paper, we focus on exploring and resolving knowledge conflicts in RALMs.
First, we present an evaluation framework for assessing knowledge conflicts across various dimensions.
% scenarios, models and evaluation 
Then, we investigate the behavior and preference of RALMs from the following two perspectives:
(1) Conflicts between internal memory and external sources: We find that capable RALMs emerge with the \textbf{Dunning-Kruger effect}, persistently favoring their incorrect internal memory even when correct evidence is provided.
Moreover, RALMs exhibit an \textbf{availability bias} towards commonly known knowledge;
(2) Conflicts between truthful, irrelevant and misleading evidence: We reveal that RALMs follow the principle of \textbf{majority rule}, preferring evidence that appears more frequently.
Furthermore, we find that RALMs exhibit \textbf{confirmation bias}, and are more willing to choose evidence that is consistent with their internal memory.
To address the issue of knowledge conflicts, we propose a method called \textbf{C}onflict-\textbf{D}isentangle \textbf{C}ontrastive \textbf{D}ecoding (\textbf{CD\textsuperscript{2}}) to better calibrate the
model's confidence.
Experimental results demonstrate that CD\textsuperscript{2} can effectively resolve knowledge conflicts.
We hope our work can provide useful insights for further research.

\section*{Limitations}
For further study, we conclude some limitations of our work as follows:

\begin{itemize}

\item We mainly investigate knowledge conflicts in RALMs from two perspectives: model performance and confidence score.
In the future, we will delve into exploring the model's mechanisms to understand the issue of knowledge conflicts.
For example, we can analyze whether knowledge conflicts will inhibit or stimulate the activation values of certain neurons.

\item In this paper, we resolve the knowledge conflicts in RALMs by calibrating the open-source model's logits.
How to mitigate the knowledge conflicts in black-box LLMs which are unable to access output logits remains to be studied.

\end{itemize}

We hope that our findings will better promote the practical application of RALMs.

\nocite{*}
\section*{Bibliographical References}\label{sec:reference}

\bibliographystyle{lrec-coling2024-natbib}
\bibliography{lrec-coling2024-example}

\section*{Language Resource References}
\label{lr:ref}
\bibliographystylelanguageresource{lrec-coling2024-natbib}
\bibliographylanguageresource{languageresource}

\end{document}